\documentclass[11pt,a4paper]{article}
\usepackage{times,latexsym}
\usepackage{url}
\usepackage[T1]{fontenc}
\usepackage{graphicx}

\usepackage{times}
\usepackage{latexsym}

\usepackage{bbding}
\usepackage{adjustbox}
\usepackage{amsmath}
\usepackage{graphicx}

\usepackage{longtable}
\usepackage{booktabs}
\usepackage{comment}
\usepackage{multirow}
\usepackage{array}
\usepackage{ulem}
\usepackage{threeparttable}
\usepackage{bbding} 

\usepackage[usenames,dvipsnames]{color}
\usepackage{colortbl}
\definecolor{mygray}{gray}{.9}

\usepackage{amsmath}
\usepackage{amssymb}
\usepackage[acceptedWithA]{tacl2021v1}

\usepackage{xspace,mfirstuc,tabulary}

\newif\iftaclinstructions
\taclinstructionsfalse 
\iftaclinstructions
\renewcommand{\confidential}{}
\renewcommand{\anonsubtext}{(No author info supplied here, for consistency with
TACL-submission anonymization requirements)}
\newcommand{\instr}
\fi

\iftaclpubformat 

\else

\fi

\title{DERA: Dense Entity Retrieval for Entity Alignment in Knowledge Graphs}

\author{
  Zhichun Wang \and Xuan Chen 
  \\
  Beijing Normal University, Beijing, China
  \\
  \texttt{zcwang@bnu.edu.cn}
}

\date{}

\begin{document}
\maketitle
\begin{abstract}
Entity Alignment (EA) aims to match equivalent entities in different Knowledge Graphs (KGs), which is essential for knowledge fusion and integration. Recently, embedding-based EA has attracted significant attention and many approaches have been proposed. Early approaches primarily focus on learning entity embeddings from the structural features of KGs, defined by relation triples. Later methods incorporated entities' names and attributes as auxiliary information to enhance embeddings for EA. However, these approaches often used different techniques to encode structural and attribute information, limiting their interaction and mutual enhancement. In this work, we propose a dense entity retrieval framework for EA, leveraging language models to uniformly encode various features of entities and facilitate nearest entity search across KGs. Alignment candidates are first generated through entity retrieval, which are subsequently reranked to determine the final alignments. We conduct comprehensive experiments on both cross-lingual and monolingual EA datasets, demonstrating that our approach achieves state-of-the-art performance compared to existing EA methods.
\end{abstract}

\section{Introduction}
Knowledge Graphs (KGs) represent structured information of entities in various domains, which facilitates machines to handle domain knowledge. Most published KGs, such as YAGO\cite{Rebele2016YAGOAM}, DBpedia\cite{bizer2009dbpedia}, and WikiData\cite{wikidata}, are heterogeneous because they are either built from different data sources or by different organizations using varying terminologies. To integrate knowledge in separate KGs, it is essential to perform Entity Alignment (EA), which aims to discover equivalent entities in different KGs. 

The problem of EA has been studied for years and many approaches have been proposed. Early EA approaches rely on manually designed features to compute similarities of entities\cite{semantic_survey}. Recently, embedding-based EA has attracted much attention, many approaches have been proposed and achieved promising performance. These approaches first embed entities in low-dimensional vector spaces, and then discover entity alignments based on distances of entity embeddings. There are mainly two paradigms of KG embedding, translation-baed methods and Graph Neural Network(GNN)-based methods. Translation-based methods learn entity embeddings using TransE or its extended models, including MTransE\cite{chen2016multi}, JAPE\cite{jape}, and BootEA\cite{bootea}, etc. GNN-based methods learn neighborhood-aware representations of entities by aggregating features of their neighbors, such approaches include GCN-Align\cite{gcnalign}, MuGNN\cite{mugnn}, and AliNet\cite{AliNet}, etc. 

Early embedding-based methods focus on structure embedding of KGs, to further improve the EA results, some latter approaches explore entities' names and attributes as side information to enhance the entity embeddings. Names and attribute values are encoded by using character or word embedding techniques, for example in MultiKE\cite{multike}, AttrGNN\cite{AttrGNN} and CEA\cite{CEA-ICDE2020}, etc. Most recently, pre-trained language models (PLMs) have also been used to encode the names and attribute values, such as in BERT-INT\cite{bertint}, SDEA\cite{SDEA}. 

Although continuous progress has been achieved in recently years, we find that there lacks a unified and effective way to encode all kinds of information of entities for EA. Most of the existing approaches encode structure information (relations) and attribute information (names, attributes, and descriptions, etc.) separately. Two kinds of information are encoded in different spaces, which are integrated before matching entities. Such EA paradigm faces both structure heterogeneity and attribute heterogeneity problems, which hinders their mutual enhancement. 

Recently, the emergence of pre-trained language models has significantly enhanced the quality of text embeddings, proving highly effective in information retrieval, question answering and retrieval-augmented language modeling. Inspired by the recent development of embedding-based IR (dense retrieval), where relevant answers to a query are retrieved based on their embedding similarities, we formalize entity alignment in KGs as an entity retrieval problem. To find equivalent entities of two KGs, entities in one KG are used as queries to retrieval the most similar entities in the other KG. In this entity retrieval framework, different kinds of entities' information can be uniformly represented in  textual forms, and we can leverage the advance of language models in embedding and searching similar entities. 

More specifically, we make the following contributions in this work:
\begin{itemize}
	\item We formalize the EA problem as an entity retrieval task, and propose a language model based framework for this task. Within this framework, entities' information are uniformly transformed into textual descriptions, which are then encoded by language model based embedding model for nearest entity search between KGs.  
	\item We propose an entity verbalization model to generate homogenous textual descriptions of entities from their heterogeneous triples. We build a synthetic triple-to-text dataset by prompting GPT, which is used for effective training the verbalization model.
	\item We design embedding models for entity retrieval and alignment reranking. The embedding model for entity retrieval encodes entities independently, which can efficiently find alignment candidates; the embedding model for alignment reranking encodes features of entity pairs, which captures the interactions of entities and guarantees the precision of alignments. 
	\item We conduct comprehensive experiments on five datasets, and compare our approach with the existing EA approaches. The results show that our approach achieves state-of-the-art results.
\end{itemize}

The rest of this paper is organized as follows:  Section 2 covers the preliminaries of our work, Section 3 details our proposed approach, Section 4 presents the experiments, Section 5 discusses related work, and Section 6 provides the conclusion.

\section{Preliminaries}
In this section, we introduce the problem of entity alignment in knowledge graphs, and formalize the task of dense entity retrieval for EA. 
  
\subsection{KG and Entity Alignment}

\noindent\textbf{Knowledge Graph (KG).} 
KGs represent structural information about entities as triples having the form of $\langle s, p, o\rangle$. A triple can be relational or attributional, a relational triple describes certain kind of relation between entities, and an attributional triple describes an attribute of an entity. In this work, we consider both relational and attributional triples in KGs. Formally, we represent a KG as $G=(E,R,A,L,T)$, where $E$, $R$, $A$ and $L$ are sets of entities, relations, attributes, and literals; $T\subseteq (E\times R\times E) \cup (E\times A\times L)$ is the sets of triples. 

\noindent\textbf{Entity Alignment (EA).}  
Given two KGs $G_s$ and $G_t$, and a set of pre-aligned entity pairs $S = \{(u,v)|u\in G_s, v\in G_t, u \equiv v\}$ ($\equiv$ denotes equivalence), the task of entity alignment is to find new equivalent entity pairs between $G_s$ and $G_t$. 

\begin{figure*}[ht]
\centering
	\includegraphics[width=1\textwidth]{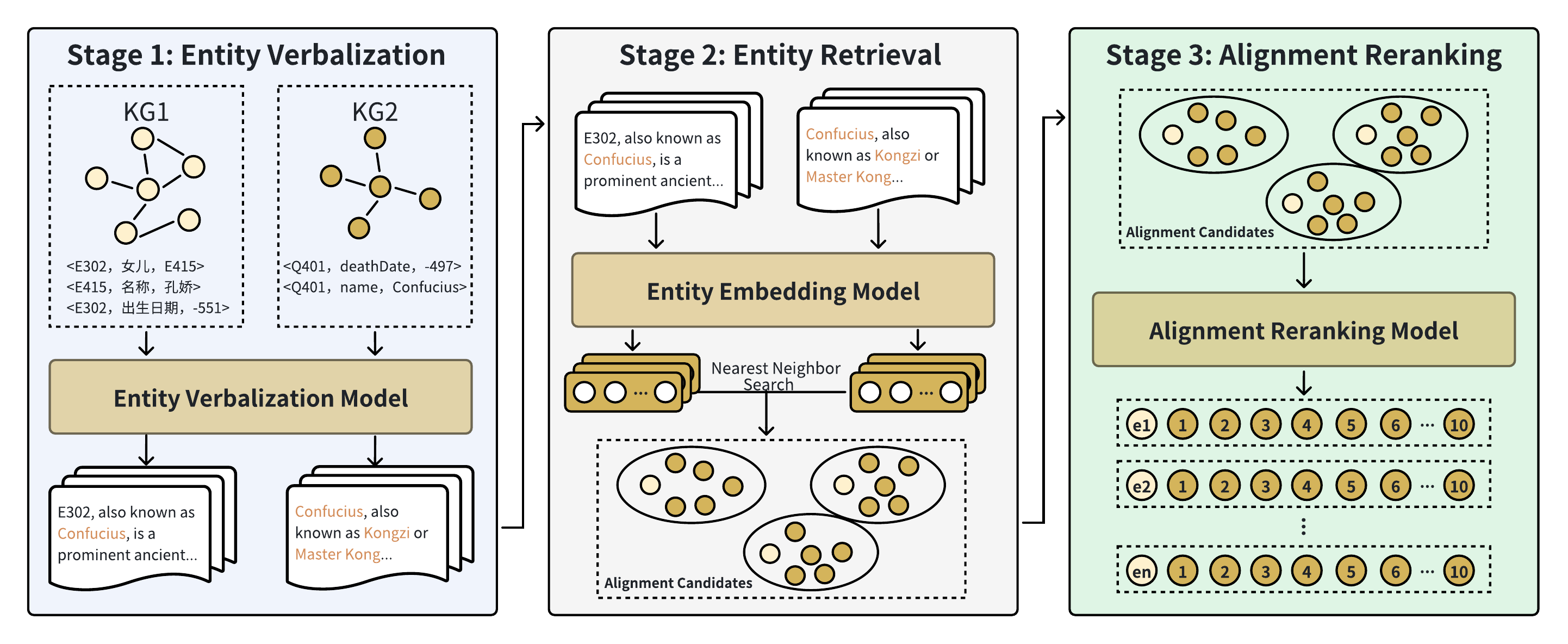}
	\caption{Framework of DERA.}
  \label{fig:framework}
\end{figure*}

\subsection{Dense Entity Retrieval}
\label{sec:problem-DER}
In this work, we formalize EA as an entity retrieval task. Given a source KG $G_s$ and a target KG $G_t$, entity retrieval aims to, for each entity $s\in G_s$, return a ranked list of k most similar entities $\left[t_1,t_2,...,t_k\right]$ in $G_t$. The top-ranked entity $t_1$ is considered as be equivalent to the source entity $s$, i.e. $s\equiv t_1$. 

To achieve accurate entity retrieval, LM-based embedding models are leveraged in our approach to encode entities into dense vectors, and the similarities of entities are computed using their vectors:

\begin{equation}
	f(s, t)=\operatorname{sim}(\phi(s), \psi(t))
\end{equation}
where $\phi(\cdot) \in \mathbb{R}^d$ and $\psi(\cdot) \in \mathbb{R}^d$ are encoders mapping the source and target entities into d-dimensional vector space, respectively. In this work, we will use the same encoder for source and target entities, and use dot-product for computing the similarity of entities. 

\section{Method}
In this section, we present the proposed EA framework \textbf{DERA} (\textbf{D}ense \textbf{E}ntity \textbf{R}etrieval for entity \textbf{A}lignment), which is shown in Figure~\ref{fig:framework}. Given two KGs to be aligned, DERA works in three main stages. \textbf{(1) Entity Verbalization (EV)}: this stage converts heterogeneous triples of entities into homogeneous natural language descriptions. Relations and attributes expressed in different languages will also be converted into one language. \textbf{(2) Entity Retrieval (ER)}: entities' textual descriptions are encoded in the same vector space. Entities are indexed using their embeddings, similar entities are retrieved based on embedding similarity to obtain alignment candidates. \textbf{(3) Alignment Reranking (AR)}: candidate alignments are further reranked by an reranking model to produce the final results.

\subsection{Entity Verbalization}
The purpose of entity verbalization is to convert relational and attribute triples of entities into textual descriptions in one language, which can be well encoded by a language model based embedding model. Given an entity $e$ in a KG, let $\mathcal{N}_{e}=\{(r_{i},e_{i})\}_{i=1}^{k}$ be the set of neighbors and associated relations of entity $e$, $\mathcal{L}_{e}=\{(a_{j},v_{j})\}_{j=1}^{m}$ be the set of attributes and values of entity $e$; here $e_i$ is an entity and $r_{i}$ is the relation connecting two entities, $v_{j}$ is the value of $a_{j}$ of $e$. Entity verbalization can be formally defined as a mapping $g\left(\mathcal{N}_e, \mathcal{L}_e\right) \rightarrow s_e$, where $s_e$ is the textual sequence of $e$.

To get high-qualified verbalization results, we train a generative language model which takes triples as input context and generate textual descriptions as outputs. More specifically, we take open Large Language Models (LLMs) as base models, and build triple-to-text dataset to fine-tune base models.

\begin{figure}
\centering
	\includegraphics[width=0.45\textwidth]{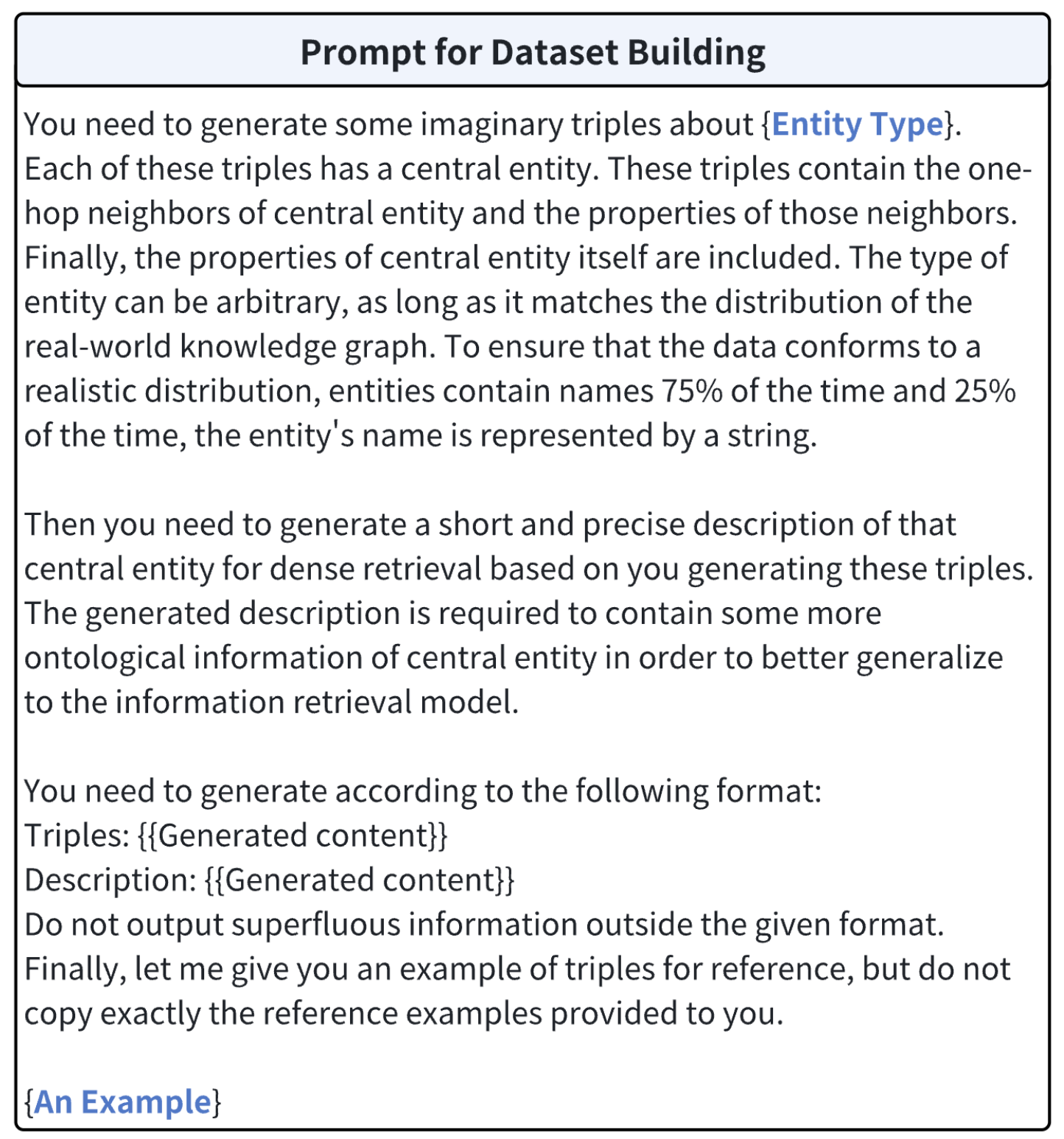}
	\caption{Prompt for Building Training Dataset for Entity Verbalization Model.}
  \label{fig:prompt}
\end{figure}

\noindent\textbf{Dataset Building.} The triple-to-text dataset is built by instructing the GPT4 using a designed prompt template, which is shown in Figure~\ref{fig:prompt}. There are four parts in the prompt: (1) The first part is an instruction prefix to describe the task of generating triples of entities of specified type; we predefined 25 common entity types, including person, organization, movie, disease, etc. (2) The second part tells the model to generate a short and precise description of the generated triples; (3) The third part specifies the formates of generated triples and textual descriptions; (4) The fourth parts gives an example to the model. 

Using the above prompt, we build a dataset contain triples and textual descriptions of 18,572 entities.

\noindent\textbf{Model Training.}
Using the generated dataset, we fine-tune the LLMs with the next word prediction task, which is a universal approach to training LLMs. For an entity, given the sequence of triples \( x = (e, r_{1},e_{1},...,r_{k},e_{k},a_{1},v_{1},...,a_{m},v_{m}) \) and its target textual description \( y = (y_1, y_2, \ldots, y_n) \), the training objective of EV model can be formulated as:
\begin{equation}
	\mathcal{L}_{\text{EV}} = -\frac{1}{n}\sum_{t=1}^n \log P(y_t | x, y_{<t}; \theta)
\end{equation}
where $n$ is the length of $y$, $y_{t} (t=1,2,...,n)$ denotes the textual tokens of the sequence $y$, $\theta$ represents the model parameters. 

In this work, we choose LLMs of 7B size as the base models of EV. More specifically, Mistral-7B-Instruct-v0.2\cite{mistral7b} and Qwen1.5-7B-Chat\cite{bai2023qwen} are used because they have excellent performances in small-size LLMs. QWen is used for EA tasks involving Chinese language, because it has great ability of handling Chinese texts. In the other EA tasks, Mistral model is used in EV stage. EV models are trained independent of specific EA tasks, once two EV models have been trained, their parameters are frozen and will not be changed in the following two stages.

\subsection{Entity Retrieval}

In this stage, entity embedding model is trained to encode entity descriptions into vector space, where entities are close to their equivalent counterparts. Using the entity embedding results, alignment candidates are produced based on embedding similarities of entities. In this work, we use a text embedding model as the basis, and fine-tune it with pre-aligned entities to further improve the embedding quality. More specifically, BGE\cite{chen2024bge} embedding model is used here because it achieves state-of-the-art performances on multilingual and cross-lingual retrieval tasks.

\noindent\textbf{Model Training.} As defined in Section~\ref{sec:problem-DER}, the similarity of two entities $s$ and $t$ is computed as the doc product of their embeddings:
\begin{equation}
	f(u, v)=\phi(u)\cdot \phi(v).
\end{equation}
Here $\phi(\cdot) \in \mathbb{R}^d$ denotes the entity embedding model which maps the entity into $d$-dimensional vector space. Given a set of seed alignments $S = \{(u,v)|u\in G_s, v\in G_t, u \equiv v\}$, the entity embedding model in our approach is trained by minimizing the following contrastive loss:
\begin{equation}
\small
	\mathcal{L}_{\text{ER}} = -\sum_{\left({u}, {v}\right)\in S} \log \frac{e^{f\left({u}, {v}\right) }}{e^{f\left({u}, {v}\right) }+\sum_{v^{\prime}\in N_u} e^{f\left({u}, v^{\prime}\right) }}
\end{equation}
where $N_u$ is a set of negative (inequivalent) entities for $u$. 

\noindent\textbf{Candidate Selection.} After the entity embedding model is trained, all the entities in two KGs can be encoded as vectors in the same space. Then candidate alignments are obtained by using each source entity to retrieval nearest target entities based on their embeddings. More specifically, for each source entity $u\in G_s$, a set of top-k nearest target entities in $G_t$ are retrieved, which are candidate alignments $u$, denoted as $V_u$. 

\subsection{Alignment Reranking}
In the entity retrieval stage, entities' descriptions are encoded independently from each other. To further improve the EA results, we design an alignment reranking model which capture the interactions of entities' features. Here a reranker built upon BERT is trained, which takes features of two entities as inputs, and predict the fine-grained similarities of entity pairs. Entity pairs are restricted to the candidates generated by the entity retrieval stage, which helps our approach to control the computation costs in alignment reranking. 

Let $C=\{(u_j,V_{u_j})\}_{j=1}^l$ be the alignment candidates, where $u_j$ is a source entity and $V_{u_j}$ is the set of its candidate equivalent entities. We construct a dataset for training our alignment reranking model, let it be $R=\{(u_j,v_j,N_j)\}_{j=1}^l$, where $(u_j,v_j)\in S$ is the pre-aligned entity pair and $N_j=V_{u_j}/\{v_j\}$ is the set of candidate entities that are not equivalent to $u_j$. The reranking model is trained by minimizing the following loss:

\begin{equation}
\small
	\mathcal{L}_{\text{AR}} = -\sum_{\left({u}_j, {v}_j,N_j\right)\in R} \log \frac{e^{\delta\left({u}_j, {v}_j\right) }}{e^{\delta\left({u}_j, {v}_j\right) }+\sum_{v^{\prime}_{k}\in N_j} e^{\delta\left({u}_j, v^{\prime}_{k}\right) }}
\end{equation}
Here $\delta\left(u,v\right)$ is the similarity score computed by the reranking model based on the inputs of two entities:
\begin{equation}
	\delta\left(u,v\right) = \operatorname{MLP}\left(\operatorname{BERT_{[CLS]}}\left(d_{u},d_{v}\right)\right)
\end{equation}
where $d_{u}$ and $d_{v}$ represent the textual descriptions of $u$ and $v$, respectively.

\begin{table*}[ht]
  \centering
  \caption{Statistics of Experimental Datasets}
  \small
  \begin{tabular}{lcrrrrrr}
    \toprule
    Dataset                           & Language & Entities & Relations & Attributes & Rel. Triples & Attr. Triples \\
    \midrule
    \multirow{2}{*}{DBP15K$_\text{ZH-EN}$} & ZH & 19,388 & 1,701 & 8,113 & 70,414  & 379,684 \\
                              & EN & 19,572 & 1,323 & 7,173 & 95,142  & 567,755 \\
\midrule
\multirow{2}{*}{DBP15K$_\text{JA-EN}$} & JA & 19,814 & 1,299 & 5,882 & 77,214  & 354,619 \\
                              & EN & 19,780 & 1,153 & 6,066 & 93,484  & 497,230 \\
\midrule
\multirow{2}{*}{DBP15K$_\text{FR-EN}$} & FR & 19,661 & 903   & 4,547 & 105,998 & 354,619 \\
                              & EN & 19,993 & 1,208 & 6,422 & 115,722 & 497,230 \\
\midrule
\multirow{2}{*}{D-W-15K-V2}   & EN & 15,000 & 167   & 175      & 73,983  & 66,813        \\
                              & EN & 15,000 & 121   & 457      & 83,365  & 175,686        \\
\midrule
\multirow{2}{*}{MED-BBK-9K}   & ZH & 9,162  & 32    & 19    & 158,357 & 11,467   \\
                              & ZH & 9,162  & 20    & 21    & 50,307  & 44,987   \\ 
    \bottomrule
  \end{tabular}
  \label{tab:ea_datasets_statistics_chap03}
\end{table*}

\begin{table*}[h]
  \centering
  \footnotesize
  \caption{Results on DBP15K Datasets}
  \resizebox*{\linewidth}{!}{
  \begin{tabular}{c|c|ccc|ccc|ccc}
    \toprule
    \multirow{2}{*}{Info.} & \multirow{2}{*}{Model}  & \multicolumn{3}{c|}{DBP15K-ZH-EN} & \multicolumn{3}{c|}{DBP15K-JA-EN} & \multicolumn{3}{c}{DBP15K-FR-EN} \\
    & & Hits@1 & Hits@10 & MRR  & Hits@1 & Hits@10 & MRR  & Hits@1 & Hits@10 & MRR  \\
    \midrule
    \multirow{4}{*}{\rotatebox[origin=c]{90}{Attributes}} & JAPE & 0.412& 0.745& 0.490& 0.363& 0.685& 0.476& 0.324& 0.667& 0.430\\
    &GCN-Align & 0.413& 0.744& 0.549& 0.399& 0.745& 0.546& 0.373& 0.745& 0.532\\
    &JarKA & \underline{0.706}& \underline{0.878}& \underline{0.766}& \underline{0.646}& \underline{0.855}& \underline{0.708}& \underline{0.704}& \underline{0.888}& \underline{0.768}\\
    &DERA(Ours)& \textbf{0.946} & \textbf{0.982} & \textbf{0.961} & \textbf{0.921}& \textbf{0.959}& \textbf{0.937}& \textbf{0.949}& \textbf{0.985}& \textbf{0.964}\\
	\midrule
    \multirow{5}{*}{\rotatebox[origin=c]{90}{Names}} & GMNN & 0.679& 0.785& --   & 0.740& 0.872& --   & 0.894& 0.952& --   \\
    & SelfKG & 0.745& 0.866& --   & 0.816& 0.913& --   & 0.957& 0.992& --   \\
    & TEA-NSP & 0.815& 0.953& 0.870 & \textbf{0.890} & \textbf{0.967} & \textbf{0.920}  & \underline{0.968} & 0.995 & \underline{0.980}  \\
    & TEA-MLM & \underline{0.831} & \underline{0.957} & \underline{0.880}  & \underline{0.883} & \underline{0.966} & \underline{0.910} & \underline{0.968}& 0.994& \underline{0.980} \\
    &  DERA(Ours) &\textbf{0.846} & \textbf{0.962}& \textbf{0.900}& 0.866& 0.951& 0.889& \textbf{0.980} & \textbf{0.996} & \textbf{0.987} \\
	\midrule
	\multirow{7}{*}{\rotatebox[origin=c]{90}{Names \& Attributes}} & HMAN & 0.871& 0.987& --   & 0.935 & \textbf{0.994} & --    & 0.973 & \underline{0.998} & --    \\
	& AttrGNN  & 0.796& 0.929& 0.845 & 0.783& 0.921& 0.834 & 0.919& 0.978& 0.910\\
    & BERT-INT & \textbf{0.968}& \underline{0.990}& \underline{0.977} & \underline{0.964} & 0.991& \underline{0.975} & \textbf{0.992}& \underline{0.998}& \textbf{0.995}\\
    & ICLEA & 0.884& 0.972& --  & 0.924& 0.978& --  & \underline{0.991} & \textbf{0.999} & --   \\
    & TEA-NSP  & \underline{0.941}& 0.983& 0.960 & 0.941 & 0.979 & 0.960 & 0.979 & 0.997 & \underline{0.990}  \\
    & TEA-MLM  & 0.935 & 0.982 & 0.950 & 0.939& 0.978& 0.950 & 0.987& 0.996& \underline{0.990} \\
    & DERA(Ours) & \textbf{0.968} & \textbf{0.994} & \textbf{0.979} & \textbf{0.967} & \underline{0.992} & \textbf{0.978} & 0.989 & \textbf{0.999} & \textbf{0.995} \\
	\midrule
	\multirow{11}{*}{\rotatebox[origin=c]{90}{Translated Names}} &  HGCN-JE  & 0.720& 0.857& --   & 0.766& 0.897& --   & 0.892& 0.961& --   \\
    & RDGCN & 0.708& 0.846& 0.746& 0.767& 0.895& 0.812& 0.886& 0.957& 0.911\\
    & NMN& 0.733& 0.869& --   & 0.785& 0.912& --   & 0.902& 0.967& --   \\
    & DERA(Ours) &{0.930} & {0.982} & {0.950} & {0.917} & {0.978} & {0.941} &{0.972} & {0.995} & {0.982} \\
    & DATTI\textsuperscript{\dag} & 0.890 & 0.958 & --    & 0.921& 0.971& --   & 0.979& 0.990& --   \\
    & SEU\textsuperscript{\dag}& 0.900& 0.965& 0.924& 0.956& 0.991& 0.969& 0.988 & \textbf{0.999} & 0.992 \\
    & EASY\textsuperscript{\dag} & 0.898 & 0.979 & 0.930 & 0.943 & 0.990 & 0.960 & 0.980 & 0.998 & 0.990 \\
    & CPL-OT\textsuperscript{\dag} & 0.927& 0.964& 0.940& 0.956& 0.983& 0.970& 0.990& 0.994& 0.990\\
    & UED\textsuperscript{\dag}& 0.915& --   & --   & 0.941& --   & --   & 0.984& --   & --   \\
    & LightEA\textsuperscript{\dag}  & \underline{0.952}& \underline{0.984}& \underline{0.964}& \underline{0.981}& \underline{0.997}& \underline{0.987}& \underline{0.995}& \underline{0.998}& \underline{0.996}\\
    & DERA\textsuperscript{\dag}(Ours)  &\textbf{0.985} & \textbf{0.997} & \textbf{0.990} & \textbf{0.994} & \textbf{0.999} & \textbf{0.996} & \textbf{0.996} & \textbf{0.999} & \textbf{0.997} \\
    \bottomrule
 \end{tabular}
}
\leftline{Approaches with $\textsuperscript{\dag}$ employ optimal transport strategy.}
\label{tab:over_performance}
\end{table*}

\section{Experiments}

\subsection{Datasets}

\textbf{Datasets.} To evaluate the performance of our approach, we conduct experiments on both cross-lingual and monolingual datasets, including: 
\begin{itemize}
	\item DBP15K\cite{jape} contains three cross-lingual EA datasets build from DBpedia, including Chinese-English (ZH-EN), Japanese-English (JA-EN), and French-English (FR-EN). 
	\item D-W-15K\cite{openea} is a monolingual EA dataset built from DBpedia and Wikipedia by using an iterative degree-based sampling method. Compared with DBP15K, D-W-15K contains KGs that are more like real-world ones.
	\item MED-BBK-9K\cite{medbbk9k} is a dataset built from two medical knowledge graphs, containing triples on diseases, symptoms, drugs, and diagnosis methods. It poses a more complex and realistic scenario for EA compared to traditional datasets like DBpedia.
\end{itemize}

Table \ref{tab:ea_datasets_statistics_chap03} shows the detailed statistics of these datasets.

\subsection{Training Details} 
We train the Entity Verbalization (EV), Entity Retrieval (ER), and Alignment Reranking (AR) models sequentially. 

\noindent\textbf{EV Model.}
In the training of EV model, we employ Deepspeed\footnote{{\url{https://github.com/microsoft/DeepSpeedExamples}}} with a context window length of 2048, the learning rate is set to $9.65e-6$, and the batch size is 24 per GPU. For the base language models, we use Qwen1.5-7B-Chat\cite{bai2023qwen} for DBP15K$_\text{ZH-EN}$ and MED-BBK-9K datasets, and use Mistral-7B-Instruct-v0.2\cite{mistral7b} for DBP15K$_\text{JA-EN}$, DBP15K$_\text{FR-EN}$, and D-W-15K datasets. Gradient accumulation is set to 1. To optimize memory usage and computation speed, we utilize Zero-Stage-3\cite{zero}, gradient checkpointing\cite{gradient_checkpointing}, and flash attention 2\cite{flashattention2}. The model is trained on 8 NVIDIA A800 GPU for 3 epochs using the AdamW optimizer.

\noindent\textbf{ER Model.} 
In the training of ER model, for each positive entity, 64 negative entities are randomly sampled from the top-200 nearest ones. The learning rate is set to $1e-5$, and the batch size to 16. We utilize distributed negative sample sharing and gradient checkpointing\cite{gradient_checkpointing}, evaluate the model every 20 steps and saving the best model based on the MRR metric on the validation set. Training is performed on 2 NVIDIA A800 GPUs for 5 epochs.

\noindent\textbf{AR Model.}
In the training of AR model, for each positive entity, 110 negative entities are randomly sampled from the top-200 nearest ones. The maximum text length is set to 512; the learning rate to $1e-5$, and the batch size to 12 per GPU. Gradient accumulation steps are set to 8. We enable gradient checkpointing, evaluate the model every 10 steps, and save the best model based on the Hits@1 metric on the validation set. Training is carried out on 2 NVIDIA A800 GPUs for 5 epochs.

\subsection{Results on DBP15K}
We compare our approach with four groups of baselines on DBP15K datasets, which are categorized by the used side information: (1) approaches using attributes as side information, including JAPE\cite{jape}, GCN-Align\cite{gcnalign}, JarKA\cite{jarka}; (2) approaches using entity names as side information, including GMNN\cite{gmnn}, SelfKG\cite{selfkg} and TEA-NSP, TEA-MLM\cite{tea}; (3) approaches using attributes and names as side information, including HMAN\cite{hman}, AttrGNN\cite{AttrGNN}, BERT-INT\cite{bertint}, ICLEA\cite{iclea} and TEA-NSP, TEA-MLM\cite{tea}; (4) approaches using translated entity names as side information, including HGCN-JE\cite{hgcn}, RDGCN\cite{rdgcn}, NMN\cite{nmn}, DATTI\cite{datti}, SEU\cite{seu}, EASY\cite{easy}, CPL\cite{cplot}, UED\cite{ued} and LigthEA\cite{lightea}. Our approach is compared to baselines in each group using the same inputs as them. Table ~\ref{tab:over_performance} outlines the results of all the approaches on DBP15K datasets. The best results in each group are highlighted in boldface, the second best results are highlighted with underlines. 

\noindent\textbf{Attributes as Side Information.} Approaches in this group align entities based on relations and attributes in KGs. Compared with approaches in this group, our approach obtains significantly better results, with average improvements of 25.3\% of Hits@1 and 20.7\% of MRR over the second best approach on three datasets. 

\noindent\textbf{Names as Side Information.} Approaches in this group use entity names and relations to discover equivalent entities. Our approach gets the best results of Hits and MRR on ZH-EN and FR-EN datasets, it obtains 1.5\% and 1.4\% improvements of Hits@1 over the second best approach TEA-MLM. While on the JA-EN dataset, TEA-NSP gets slightly better results than ours. 

\noindent\textbf{Names and Attributes as Side Information.} When using both names and attributes, our approach still obtain top-ranked results. Except for the Hits@10 on JA-EN and Hits@1 on FR-EN datasets, our approach gets the best results among all the compared approaches in this group.  

\noindent\textbf{Translated Names as Side Information.} Approaches in this group use machine translation tool to convert non-English names into English ones, and takes translated names as side information. Some of the approaches (annotated with $\textsuperscript{\dag}$) in this group also employ optimal transport strategies to draw final alignments from entity similarities, which can effectively promote the results. To be fairly compared with these approach, we also report the results of our approach with the optimal transport strategy. According to the results, our approach gets the best results among all the approaches in this group. Among approaches without optimal transport strategies, our approach also gets the best results.  

\subsection{Results of Hard Setting on DBP15K}
In the work of AttrGNN\cite{AttrGNN}, a hard setting of evaluations on DBP15K was proposed. The purpose of this hard setting is to build more difficult testing set on DBP15K. Specifically, similarities of equivalent entities in the datasets are first measured using embeddings of their names, 60\% entity pairs with the lowest similarities are selected as the testing set, and the remaining entity pairs are randomly split into training set (30\%) and validation set (10\%). 

\begin{figure*}[ht]
 \centering
 \includegraphics[width=0.8\linewidth]{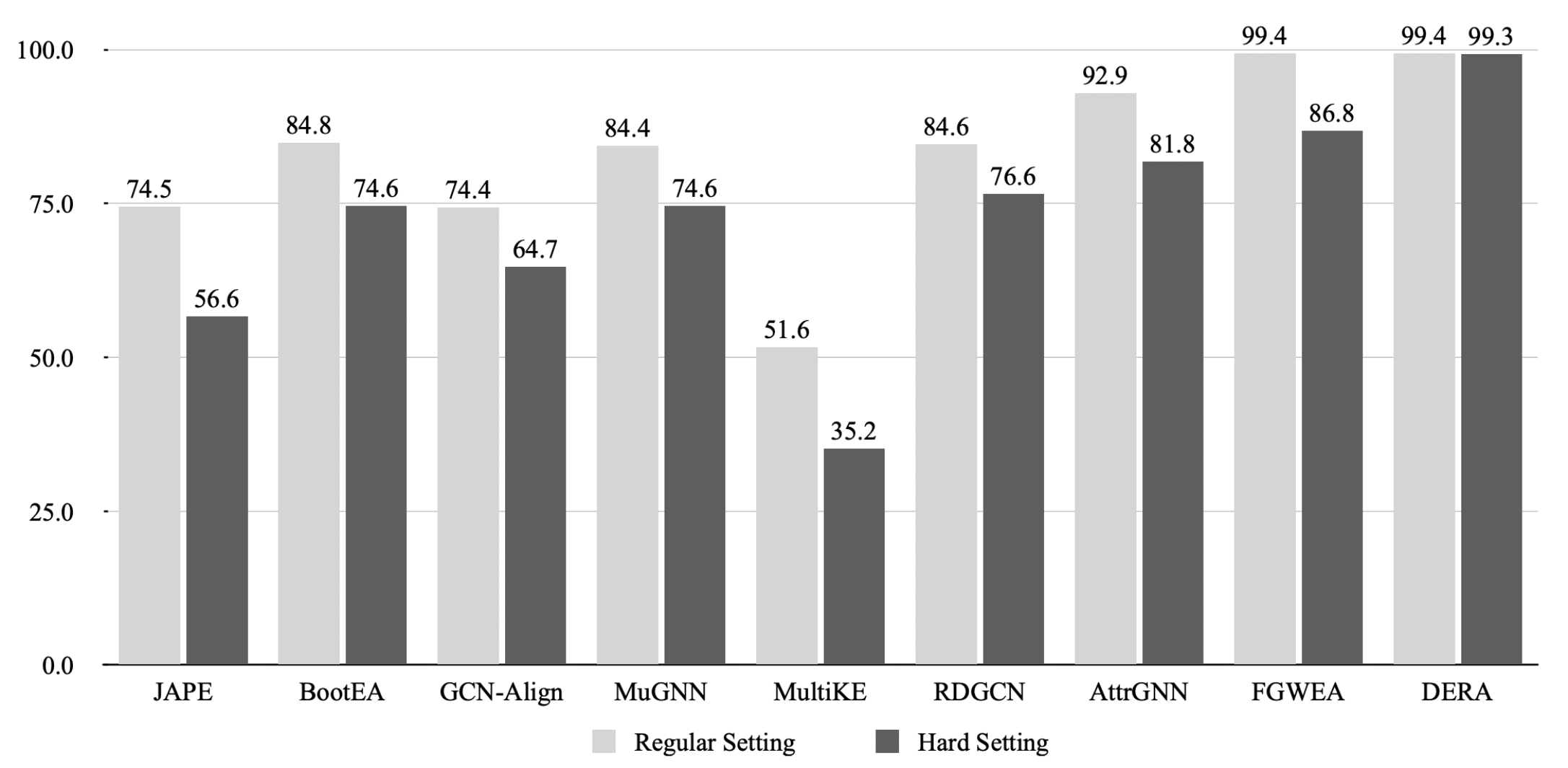}
 \caption{Hits@10 (\%) of approaches under the regular setting and the hard setting on DBP15k.}
 \label{fig:hard_regular_setting_drop}
\end{figure*}

Table~\ref{tab:hard_setting} shows that results of hard setting on DBP15K. Our approach is compared with eight baselines, including JAPE \cite{jape}, BootEA\cite{bootea}, GCN-Align\cite{gcnalign}, MuGNN\cite{mugnn}, MultiKE\cite{multike}, RDGCN\cite{rdgcn}, AttrGNN\cite{AttrGNN}, and FGWEA\cite{fgwea}. According to the results, our approach DERA gets the best Hits and MRR on all of the three datasets. Figure ~\ref{fig:hard_regular_setting_drop} compares the Hits@10 of approaches in regular setting and hard setting on DBP15K$_\text{ZH\_EN}$. All the baselines have significant decreases of Hits@10, while DERA (using names and attributes as side information) has only 0.1\% decrease, showing its remarkable robustness.

\begin{table*}[ht]
    \centering
    \caption{Results of Hard Setting on DBP15K}
	\small
    \begin{tabular}{c|ccc|ccc|ccc}
      \toprule
    \multirow{2}{*}{Model} & \multicolumn{3}{c|}{DBP15K-ZH-EN}  & \multicolumn{3}{c|}{DBP15K-JA-EN}  & \multicolumn{3}{c}{DBP15K-FR-EN}  \\
    & Hits1 & Hits10      & MRR   & Hits1 & Hits10      & MRR   & Hits1 & Hits10      & MRR   \\
    \midrule
    JAPE       & 0.350 & 0.566 & 0.451 & 0.311 & 0.520 & 0.410 & 0.253 & 0.483 & 0.361 \\
    BootEA    & 0.513 & 0.746 & 0.593 & 0.493 & 0.746 & 0.578 & 0.513 & 0.769 & 0.603 \\
    GCN-Align  & 0.366 & 0.647 & 0.464 & 0.339 & 0.653 & 0.448 & 0.303 & 0.637 & 0.414 \\
    MuGNN      & 0.406 & 0.746 & 0.521 & 0.399 & 0.753 & 0.515 & 0.407 & 0.783 & 0.531 \\
    MultiKE    & 0.279 & 0.352 & 0.306 & 0.482 & 0.557 & 0.509 & 0.647 & 0.695 & 0.665 \\
    RDGCN      & 0.604 & 0.766 & 0.662 & 0.682 & 0.838 & 0.737 & 0.829 & 0.931 & 0.866 \\
    AttrGNN    & 0.662 & 0.818 & 0.719 & 0.774 & \underline{0.903} & 0.821 & 0.886 & 0.956 & 0.912 \\
    FGWEA     & \underline{0.756} & \underline{0.868} & \underline{0.796} & \underline{0.788} & 0.897 & \underline{0.828} & \underline{0.983} & \underline{0.997} & \underline{0.988} \\
    DERA(Ours)  & \textbf{0.967} & \textbf{0.993} & \textbf{0.977} & \textbf{0.959} & \textbf{0.992} & \textbf{0.973} &\textbf{0.987}       & \textbf{1.000}      &\textbf{0.993} \\      
    \bottomrule
    \end{tabular}
    \label{tab:hard_setting}
  \end{table*}

\begin{table*}
  \centering
  \caption{Results on DW-15K and MED-BBK-9K Datasets}
  \small
  \begin{tabular}{c|ccc|ccc}
 \toprule
  \multirow{2}{*}{Model} & \multicolumn{3}{c|}{DW-15K-V2} & \multicolumn{3}{c}{MED-BBK-9K} \\
  & Precision & Recall & F1 & Precision & Recall & F1 \\
  \midrule
  LogMap & -- & -- & -- & \underline{86.4} & 44.1 & 58.4 \\
  PARIS & 95.0 & 85.0 & 89.7 & 77.9 & 36.7 & 49.9 \\
  PRASE & 94.8 & 90.0 & 92.3 & 83.7 & 61.9 & 71.1 \\
  MultiKE & 49.5 & 49.5 & 49.5 & 41.0 & 41.0 & 41.0 \\
  BootEA  & 82.1 & 82.1 & 82.1 & 30.7 & 30.7 & 30.7 \\
  RSNs & 72.3 & 72.3 & 72.3 & 19.5 & 19.5 & 19.5 \\
  FGWEA\textsuperscript{\dag} & \underline{95.2} & \underline{90.3} & \underline{92.7} & \textbf{93.9} & \underline{73.2} & \underline{82.3} \\
  DERA\textsuperscript{\dag}(Ours) & \textbf{98.2} & \textbf{98.2} & \textbf{98.2} & 84.1 & \textbf{84.1} & \textbf{84.1} \\
  \bottomrule
  \end{tabular}
  \label{tab:dw15k_med}
\end{table*}

\subsection{Results on DW-15K and MED-BBK-9K} 

DW15K and MED-BBK-9K are two challenging datasets of entity alignment. DW-15K is built from Wikipedia, where entity names are replaced with ids; there are also significant missing and corrupted attribute values. The dataset of MED-BBK-9K is built from an authoritative medical KG and a KG built from a Chinese online encyclopedia (Baidu Baike); many entities in MED-BBK-9K lack names and attributes, which makes the EA task more difficult. We compared our approach with seven approaches, three of them are probabilistic ones including LogMap\cite{logmap}, PARIS\cite{paris}, and PRASE\cite{prase}; four of them are embedding-based ones including MultiKE\cite{multike}, BootEA\cite{bootea}, RSNs\cite{rsn4ea} and FGWEA\cite{fgwea}. Following the same evaluation settings of SOTA approaches on these two datasets, we report the Precision, Recall and F1 of all the compared approaches.

Table~\ref{tab:dw15k_med} outlines the results. Our approach DERA obtains 98.2\% and 84.1\% F1 scores on D-W-15K-V2 and MED-BBK-9K, respectively. Compared to the former best approach FGWEA, DERA gets 5.5\% and 1.8\% improvements of F1 scores on two datasets, respectively. It demonstrates DERA's superior performances on difficult EA tasks.

\begin{table*}
  \caption{Results of Ablation Study}
  \small
  \renewcommand{\arraystretch}{1.3}
  \resizebox{\linewidth}{!}{
  \begin{tabular}{c|ccc|ll|ll|ll}
    \toprule
  &\multirow{2}{*}{\textbf{EV}} & \multirow{2}{*}{\textbf{ER}} & \multirow{2}{*}{\textbf{AR}} & \multicolumn{2}{c|}{DBP15K-ZH-EN} & \multicolumn{2}{c|}{DBP15K-JA-EN}      & \multicolumn{2}{c}{DBP15K-FR-EN}      \\
  & &      &   & Hits@1 & MRR & Hits@1 &  MRR & Hits@1 & MRR \\
  \midrule
  \multirow{4}{*}{\rotatebox[origin=c]{90}{Attributes}} & \checkmark & \checkmark    & \checkmark & \textbf{0.946}  & \textbf{0.962} & \textbf{0.923}  & \textbf{0.940} & \textbf{0.949}  & \textbf{0.963} \\
  & $\times$ & \checkmark    & \checkmark & 0.926~$_{0.020\downarrow}$  & 0.943~$_{0.019\downarrow}$ & 0.914~$_{0.009\downarrow}$  & 0.928~$_{0.012\downarrow}$ & 0.931~$_{0.018\downarrow}$  & 0.948~$_{0.015\downarrow}$ \\
  & \checkmark & \checkmark    &  $\times$ & 0.927~$_{0.019\downarrow}$  & 0.948~$_{0.014\downarrow}$ & 0.903~$_{0.020\downarrow}$  & 0.924~$_{0.016\downarrow}$ & 0.948~$_{0.001\downarrow}$ & \textbf{0.963}~$_{0.000-}$ \\
  & $\times$ & \checkmark    &  $\times$ & 0.892~$_{0.054\downarrow}$  & 0.918~$_{0.044\downarrow}$ & 0.859~$_{0.064\downarrow}$  & 0.885~$_{0.055\downarrow}$ & 0.927~$_{0.022\downarrow}$  & 0.946~$_{0.017\downarrow}$ \\ 
  \midrule
  \multirow{4}{*}{\rotatebox[origin=c]{90}{Attrs.\& Names}} & \checkmark & \checkmark    & \checkmark & \textbf{0.968}  & \textbf{0.979} & \textbf{0.967}  & \textbf{0.978} & 0.989  & 0.994 \\
  & $\times$ & \checkmark    & \checkmark & 0.926~$_{0.042\downarrow}$  & 0.945~$_{0.034\downarrow}$ & 0.909~$_{0.058\downarrow}$  & 0.923~$_{0.055\downarrow}$ & 0.980~$_{0.009\downarrow}$  & 0.988~$_{0.006\downarrow}$ \\
  & \checkmark & \checkmark    &  $\times$ & 0.955~$_{0.013\downarrow}$  & 0.970~$_{0.009\downarrow}$ & 0.950~$_{0.017\downarrow}$  & 0.965~$_{0.013\downarrow}$ & \textbf{0.991}~$_{0.002\uparrow}$  & \textbf{0.995}~$_{0.001\uparrow}$ \\
  & $\times$ & \checkmark    &  $\times$ & 0.883~$_{0.085\downarrow}$  & 0.911~$_{0.068\downarrow}$ & 0.812~$_{0.155\downarrow}$  & 0.848~$_{0.130\downarrow}$ & 0.980~$_{0.009\downarrow}$  & 0.988~$_{0.006\downarrow}$ \\
    \bottomrule
  \end{tabular}
 }
  \label{tab:attr_only_ablation}
  \end{table*}

\subsection{Ablation Study}
To analyze the effectiveness and contribution of each component in the proposed approach, we conduct ablation studies on DBP15K datasets. We ran two groups of experiments, one group uses attributes as side information, and the other group uses both names and attributes as side information. In each group, we ran three variations of DERA: 1) DEAR without EV module, triples of entities are directly serialized to generate inputs of ER module; 2) DERA without AR module, the final alignments are returned based on the similarities computed by ER module; 3) DERA without EV and AR module. The results of the above variations of DERA are compared to the original version of DERA, changes in results are shown in small numbers after the results. 

Table ~\ref{tab:attr_only_ablation} shows the results of ablation study. According to the results, we have the following observations: 

(1) Removing the EV module in DERA leads to average 1.6\% decrease of Hits@1 and 1.5\% decrease of MRR when using attributes as side information. The average decreases become 3.6\% of Hits@1 and 3.2\% of MRR when using attributes and names as side information. It shows that EV module has positive effects on the EA results. The performance decreases more on ZH-EN and JA-EN datasets, where the involving languages are more different than the FR-EN dataset. It indicates that EV module is important in handling language heterogeneity in EA tasks. 

(2) Removing the ER module in DERA leads to average 1.3\% decrease of Hits@1 and 1.0\% of MRR when using attributes on three datasets. If attributes and names are all used as side information, DERA without AR module gets 1.5\% decrease of Hits@1 and 1.1\% decrease of MRR on ZH-EN and JA-EN datasets, 0.2\% and 0.1\% improvements of Hits@1 and MRR on FR-EN dataset. It shows that AR module works more effectively on EA tasks with high heterogeneity and linguistic differences. When the alignment results are already good enough (e.g. >99\% Hits@1 on FR-EN dataset), it is difficult for AR module to further improve the results. 

(3) Removing both AR and RR modules in DERA leads to significant performance drops on all the datasets, there are average 4.7\% decrease of Hits@1 and 3.9\% decrease of MRR when attributes are used as side information. The decreases become 8.3\% of Hits@1 and 6.8\% of MRR when attributes and names are used. Comparing with DERA variation with EV and ER module, DERA also have significant performance drops, which shows that EV module is necessary for obtaining good results.
  
\section{Related Work}

\subsection{Embedding-based EA}
Embedding-based KG alignment approaches employ TransE and GNN to learn entities' embeddings, and then find equivalent entities in the vector spaces. Early approaches mainly rely on the structure information in KGs to find alignments, including TransE-based approaches MTransE~\cite{chen2016multi}, IPTransE~\cite{zhu2017iterative}, BootEA~\cite{bootea}, etc, and GNN-based approaches MuGNN~\cite{mugnn}, NAEA~\cite{naea2019}, RDGCN~\cite{rdgcn} and AliNet~\cite{AliNet}, etc. To get improved results, some approaches utilize entity attributes or names in KGs. JAPE~\cite{jape} performs attribute embedding by Skip-Gram model which captures the correlations of attributes in KGs. GCN-Align~\cite{gcnalign} encodes attribute information of entities into their embeddings by using GCNs. MultiKE~\cite{multike} uses a framework unifying the views of entity names, relations and attributes to learn embeddings for aligning entities. CEA~\cite{CEA-ICDE2020} combines structural, semantic and string features of entities, which are integrated with dynamically assigned weights.

\subsection{Language Model-based EA}
As Pre-trained Language Models(PLMs) being successfully used in various tasks, some approaches utilize PLMs to model the semantic information of entities in the task of KG alignment.
AttrGNN\cite{AttrGNN} uses BERT to encode attribute features of entities. It encode each attribute and value separately, and then uses a graph attention network to compute the weighted average of attributes and values. 
BERT-INT\cite{bertint} embeds names, descriptions, attributes and values of entities using a LM; pair-wise neighbor-view and attribute-view interactions are performed to get the matching score of entities. The interactions are time-consuming, thus BERT-INT cannot scale to large KGs. 
SDEA\cite{SDEA} find-tunes BERT to encode attribute values of an entity into attribute embedding; attribute embeddings of neighbors are fed to BiGRU to get relation embedding of an entity. 
TEA\cite{tea} sorts triples in alphabetical order by relations and attributes to form sequences, and uses a textual entailment framework for entity alignment. TEA takes entity-pair sequence as the input of PLM, and let the PLM to predict the probability of entailment. It takes pairwise input, cannot scale to large KGs.
AutoAlign\cite{autoalign} gets attribute character embeddings and predicate-proximity-graph embeddings by using large language models.
AttrGNN, BERT-INT and SDEA use BERT to encode attribute information of entities, and then employ GNNs to incorporate relation information into entities' embeddings. Being different from these approaches, our approach directly use language model to encode both the attributes and relations of entities. TEA uses similar way to encode attribute and relation information, but it takes entity pair as input, which cannot scale to large-scale KG alignment tasks. 

As the advent of Large Language Models (LLMs), there are several approaches exploring LLMs for EA. LLMEA\cite{LLMEA} fuses the knowledge from KGs and LLMs to predict entity alignments. It first uses RAGAT to learn entity embeddings which are used to draws alignment candidates; it then uses candidate alignments as options to generates multi-choice questions, which are passed to LLMs to predict the answer. 
ChatEA\cite{ChatEA} first uses Simple-HHEA\cite{simple-hhea} to obtain candidate alignments, and then leverages LLMs' reasoning abilities to predict the final results. 
LLMEA and ChatEA all explore the reasoning abilities of LLM to predict entity alignments. Because the number of potential alignments are usually huge, they use exiting EA methods to generate alignment candidates, from which LLMs are used to select the final results. According to the results, the improvements contributed by LLMs are restricted. 
  
\section{Conclusion}
In this paper, we propose a dense entity retrieval approach, DERA, for entity alignment in knowledge graphs. DERA first converts entity triples into unified textual descriptions using an entity verbalization model, and then trains a language model-based embedding model to encode the entities. Candidate alignments are identified based on their similarities in the embedding space and are further reranked by an alignment reranking model. Experiments demonstrate that DERA achieves state-of-the-art results on entity alignment tasks of varying difficulty levels. 
  
 \section*{Limitations}
  The primary limitation of DERA is its pipelined framework, where models in its three stages are trained sequentially. Consequently, the component models in DERA are not optimized jointly during training. Exploring efficient methods for the joint learning of these models would be a valuable direction for future work, potentially enhancing the results further. Additionally, DERA consumes more GPU power than traditional models, which is another limitation.

\bibliography{dera_arxiv}

\begin{thebibliography}{49}
\expandafter\ifx\csname natexlab\endcsname\relax\def\natexlab#1{#1}\fi

\bibitem[{Bai et~al.(2023)Bai, Bai, Chu, Cui, Dang, Deng, Fan, Ge, Han, Huang,
  Hui, Ji, Li, Lin, Lin, Liu, Liu, Lu, Lu, Ma, Men, Ren, Ren, Tan, Tan, Tu,
  Wang, Wang, Wang, Wu, Xu, Xu, Yang, Yang, Yang, Yang, Yao, Yu, Yuan, Yuan,
  Zhang, Zhang, Zhang, Zhang, Zhou, Zhou, Zhou, and Zhu}]{bai2023qwen}
Jinze Bai, Shuai Bai, Yunfei Chu, Zeyu Cui, Kai Dang, Xiaodong Deng, Yang Fan,
  Wenbin Ge, Yu~Han, Fei Huang, Binyuan Hui, Luo Ji, Mei Li, Junyang Lin, Runji
  Lin, Dayiheng Liu, Gao Liu, Chengqiang Lu, Keming Lu, Jianxin Ma, Rui Men,
  Xingzhang Ren, Xuancheng Ren, Chuanqi Tan, Sinan Tan, Jianhong Tu, Peng Wang,
  Shijie Wang, Wei Wang, Shengguang Wu, Benfeng Xu, Jin Xu, An~Yang, Hao Yang,
  Jian Yang, Shusheng Yang, Yang Yao, Bowen Yu, Hongyi Yuan, Zheng Yuan,
  Jianwei Zhang, Xingxuan Zhang, Yichang Zhang, Zhenru Zhang, Chang Zhou,
  Jingren Zhou, Xiaohuan Zhou, and Tianhang Zhu. 2023.
\newblock \href {http://arxiv.org/abs/2309.16609} {Qwen technical report}.

\bibitem[{Bizer et~al.(2009)Bizer, Lehmann, Kobilarov, Auer, Becker, Cyganiak,
  and Hellmann}]{bizer2009dbpedia}
Christian Bizer, Jens Lehmann, Georgi Kobilarov, S{\"o}ren Auer, Christian
  Becker, Richard Cyganiak, and Sebastian Hellmann. 2009.
\newblock Dbpedia-a crystallization point for the web of data.
\newblock \emph{Web Semantics: science, services and agents on the world wide
  web}, 7(3):154--165.

\bibitem[{Cao et~al.(2019)Cao, Liu, Li, Liu, Li, and Chua}]{mugnn}
Yixin Cao, Zhiyuan Liu, Chengjiang Li, Zhiyuan Liu, Juanzi Li, and Tat-Seng
  Chua. 2019.
\newblock \href {https://doi.org/10.18653/v1/P19-1140} {Multi-channel graph
  neural network for entity alignment}.
\newblock In \emph{Proceedings of the 57th Annual Meeting of the Association
  for Computational Linguistics}, pages 1452--1461, Florence, Italy.
  Association for Computational Linguistics.

\bibitem[{Chen et~al.(2020)Chen, Zhang, Tang, Chen, and Li}]{jarka}
Bo~Chen, Jing Zhang, Xiaobin Tang, Hong Chen, and Cuiping Li. 2020.
\newblock Jarka: Modeling attribute interactions for cross-lingual knowledge
  alignment.
\newblock In \emph{Advances in Knowledge Discovery and Data Mining}, pages
  845--856, Cham. Springer International Publishing.

\bibitem[{Chen et~al.(2024)Chen, Xiao, Zhang, Luo, Lian, and Liu}]{chen2024bge}
Jianlv Chen, Shitao Xiao, Peitian Zhang, Kun Luo, Defu Lian, and Zheng Liu.
  2024.
\newblock \href {http://arxiv.org/abs/2402.03216} {Bge m3-embedding:
  Multi-lingual, multi-functionality, multi-granularity text embeddings through
  self-knowledge distillation}.

\bibitem[{Chen et~al.(2017)Chen, Tian, Yang, and Zaniolo}]{chen2016multi}
Muhao Chen, Yingtao Tian, Mohan Yang, and Carlo Zaniolo. 2017.
\newblock Multilingual knowledge graph embeddings for cross-lingual knowledge
  alignment.
\newblock In \emph{Proceedings of the Twenty-Sixth International Joint
  Conference on Artificial Intelligence (AAAI2017)}, pages 1511--1517.

\bibitem[{Chen et~al.(2016)Chen, Xu, Zhang, and
  Guestrin}]{gradient_checkpointing}
Tianqi Chen, Bing Xu, Chiyuan Zhang, and Carlos Guestrin. 2016.
\newblock \href {http://arxiv.org/abs/1604.06174} {Training deep nets with
  sublinear memory cost}.

\bibitem[{Dao(2023)}]{flashattention2}
Tri Dao. 2023.
\newblock Flashattention-2: Faster attention with better parallelism and work
  partitioning.
\newblock \emph{arXiv preprint arXiv:2307.08691}.

\bibitem[{Ding et~al.(2022)Ding, Zhang, and Yin}]{cplot}
Qijie Ding, Daokun Zhang, and Jie Yin. 2022.
\newblock \href {https://doi.org/10.1109/ICDM54844.2022.00107} {Conflict-aware
  pseudo labeling via optimal transport for entity alignment}.
\newblock In \emph{2022 IEEE International Conference on Data Mining (ICDM)},
  pages 915--920.

\bibitem[{Ge et~al.(2021)Ge, Liu, Chen, Zheng, and Gao}]{easy}
Congcong Ge, Xiaoze Liu, Lu~Chen, Baihua Zheng, and Yunjun Gao. 2021.
\newblock \href {https://doi.org/10.1145/3404835.3462870} {Make it easy: An
  effective end-to-end entity alignment framework}.
\newblock In \emph{Proceedings of the 44th International ACM SIGIR Conference
  on Research and Development in Information Retrieval}, SIGIR '21, pages
  777--786, New York, NY, USA. Association for Computing Machinery.

\bibitem[{Guo et~al.(2019)Guo, Sun, and Hu}]{rsn4ea}
Lingbing Guo, Zequn Sun, and Wei Hu. 2019.
\newblock \href {https://proceedings.mlr.press/v97/guo19c.html} {Learning to
  exploit long-term relational dependencies in knowledge graphs}.
\newblock In \emph{Proceedings of the 36th International Conference on Machine
  Learning}, volume~97 of \emph{Proceedings of Machine Learning Research},
  pages 2505--2514. PMLR.

\bibitem[{Jiang et~al.(2023)Jiang, Sablayrolles, Mensch, Bamford, Chaplot,
  de~las Casas, Bressand, Lengyel, Lample, Saulnier, Lavaud, Lachaux, Stock,
  Scao, Lavril, Wang, Lacroix, and Sayed}]{mistral7b}
Albert~Q. Jiang, Alexandre Sablayrolles, Arthur Mensch, Chris Bamford,
  Devendra~Singh Chaplot, Diego de~las Casas, Florian Bressand, Gianna Lengyel,
  Guillaume Lample, Lucile Saulnier, Lélio~Renard Lavaud, Marie-Anne Lachaux,
  Pierre Stock, Teven~Le Scao, Thibaut Lavril, Thomas Wang, Timothée Lacroix,
  and William~El Sayed. 2023.
\newblock \href {http://arxiv.org/abs/2310.06825} {Mistral 7b}.

\bibitem[{Jiang et~al.(2024{\natexlab{a}})Jiang, Shen, Shi, Xu, Li, Li, Guo,
  Shen, and Wang}]{ChatEA}
Xuhui Jiang, Yinghan Shen, Zhichao Shi, Chengjin Xu, Wei Li, Zixuan Li, Jian
  Guo, Huawei Shen, and Yuanzhuo Wang. 2024{\natexlab{a}}.
\newblock \href {http://arxiv.org/abs/2402.15048} {Unlocking the power of large
  language models for entity alignment}.

\bibitem[{Jiang et~al.(2024{\natexlab{b}})Jiang, Xu, Shen, Wang, Su, Shi, Sun,
  Li, Guo, and Shen}]{simple-hhea}
Xuhui Jiang, Chengjin Xu, Yinghan Shen, Yuanzhuo Wang, Fenglong Su, Zhichao
  Shi, Fei Sun, Zixuan Li, Jian Guo, and Huawei Shen. 2024{\natexlab{b}}.
\newblock \href {https://doi.org/10.1145/3589334.3645720} {Toward practical
  entity alignment method design: Insights from new highly heterogeneous
  knowledge graph datasets}.
\newblock In \emph{Proceedings of the ACM on Web Conference 2024}, WWW '24,
  pages 2325--2336, New York, NY, USA. Association for Computing Machinery.

\bibitem[{Jim{\'e}nez-Ruiz and Cuenca~Grau(2011)}]{logmap}
Ernesto Jim{\'e}nez-Ruiz and Bernardo Cuenca~Grau. 2011.
\newblock \href {https://doi.org/10.1007/978-3-642-25073-6_18} {Logmap:
  Logic-based and scalable ontology matching}.
\newblock In \emph{Proceedings of the Tenth International Semantic Web
  Conference(ISWC2011)}, pages 273--288.

\bibitem[{Liu et~al.(2022)Liu, Hong, Wang, Chen, Kharlamov, Dong, and
  Tang}]{selfkg}
Xiao Liu, Haoyun Hong, Xinghao Wang, Zeyi Chen, Evgeny Kharlamov, Yuxiao Dong,
  and Jie Tang. 2022.
\newblock \href {https://doi.org/10.1145/3485447.3511945} {Selfkg:
  Self-supervised entity alignment in knowledge graphs}.
\newblock In \emph{Proceedings of the ACM Web Conference 2022}, WWW '22, pages
  860--870, New York, NY, USA. Association for Computing Machinery.

\bibitem[{Liu et~al.(2020)Liu, Cao, Pan, Li, Liu, and Chua}]{AttrGNN}
Zhiyuan Liu, Yixin Cao, Liangming Pan, Juanzi Li, Zhiyuan Liu, and Tat-Seng
  Chua. 2020.
\newblock \href {https://doi.org/10.18653/v1/2020.emnlp-main.515} {Exploring
  and evaluating attributes, values, and structures for entity alignment}.
\newblock In \emph{Proceedings of the 2020 Conference on Empirical Methods in
  Natural Language Processing (EMNLP)}, pages 6355--6364, Online. Association
  for Computational Linguistics.

\bibitem[{Luo and Yu(2022)}]{ued}
Shengxuan Luo and Sheng Yu. 2022.
\newblock \href {https://doi.org/10.18653/v1/2022.findings-acl.183} {An
  accurate unsupervised method for joint entity alignment and dangling entity
  detection}.
\newblock In \emph{Findings of the Association for Computational Linguistics:
  ACL 2022}, pages 2330--2339, Dublin, Ireland. Association for Computational
  Linguistics.

\bibitem[{Mao et~al.(2022{\natexlab{a}})Mao, Ma, Yuan, Zhu, Wang, Xie, Wu, and
  Lan}]{datti}
Xin Mao, Meirong Ma, Hao Yuan, Jianchao Zhu, ZongYu Wang, Rui Xie, Wei Wu, and
  Man Lan. 2022{\natexlab{a}}.
\newblock \href {https://doi.org/10.18653/v1/2022.acl-long.405} {An effective
  and efficient entity alignment decoding algorithm via third-order tensor
  isomorphism}.
\newblock In \emph{Proceedings of the 60th Annual Meeting of the Association
  for Computational Linguistics (Volume 1: Long Papers)}, pages 5888--5898,
  Dublin, Ireland. Association for Computational Linguistics.

\bibitem[{Mao et~al.(2021)Mao, Wang, Wu, and Lan}]{seu}
Xin Mao, Wenting Wang, Yuanbin Wu, and Man Lan. 2021.
\newblock \href {https://doi.org/10.18653/v1/2021.emnlp-main.226} {From
  alignment to assignment: Frustratingly simple unsupervised entity alignment}.
\newblock In \emph{Proceedings of the 2021 Conference on Empirical Methods in
  Natural Language Processing}, pages 2843--2853, Online and Punta Cana,
  Dominican Republic. Association for Computational Linguistics.

\bibitem[{Mao et~al.(2022{\natexlab{b}})Mao, Wang, Wu, and Lan}]{lightea}
Xin Mao, Wenting Wang, Yuanbin Wu, and Man Lan. 2022{\natexlab{b}}.
\newblock \href {https://doi.org/10.18653/v1/2022.emnlp-main.52} {{L}ight{EA}:
  A scalable, robust, and interpretable entity alignment framework via
  three-view label propagation}.
\newblock In \emph{Proceedings of the 2022 Conference on Empirical Methods in
  Natural Language Processing}, pages 825--838, Abu Dhabi, United Arab
  Emirates. Association for Computational Linguistics.

\bibitem[{Noy et~al.(2017)Noy, Nentwig, Hartung, Ngonga~Ngomo, and
  Rahm}]{semantic_survey}
Natasha Noy, Markus Nentwig, Michael Hartung, Axel-Cyrille Ngonga~Ngomo, and
  Erhard Rahm. 2017.
\newblock \href {https://doi.org/10.3233/SW-150210} {A survey of current link
  discovery frameworks}.
\newblock \emph{Semantic Web}, 8(3):419--436.

\bibitem[{Qi et~al.(2021)Qi, Zhang, Chen, Chen, Xiang, Zhang, and
  Zheng}]{prase}
Zhiyuan Qi, Ziheng Zhang, Jiaoyan Chen, Xi~Chen, Yuejia Xiang, Ningyu Zhang,
  and Yefeng Zheng. 2021.
\newblock \href {https://doi.org/10.24963/ijcai.2021/278} {Unsupervised
  knowledge graph alignment by probabilistic reasoning and semantic embedding}.
\newblock In \emph{Proceedings of the Thirtieth International Joint Conference
  on Artificial Intelligence, {IJCAI-21}}, pages 2019--2025. International
  Joint Conferences on Artificial Intelligence Organization.
\newblock Main Track.

\bibitem[{Rajbhandari et~al.(2020)Rajbhandari, Rasley, Ruwase, and He}]{zero}
Samyam Rajbhandari, Jeff Rasley, Olatunji Ruwase, and Yuxiong He. 2020.
\newblock Zero: Memory optimizations toward training trillion parameter models.
\newblock In \emph{SC20: International Conference for High Performance
  Computing, Networking, Storage and Analysis}, pages 1--16. IEEE.

\bibitem[{Rebele et~al.(2016)Rebele, Suchanek, Hoffart, Biega, Kuzey, and
  Weikum}]{Rebele2016YAGOAM}
Thomas Rebele, Fabian~M. Suchanek, Johannes Hoffart, Joanna~Asia Biega, Erdal
  Kuzey, and Gerhard Weikum. 2016.
\newblock \href {https://doi.org/10.1007/978-3-319-46547-0_19} {Yago: A
  multilingual knowledge base from wikipedia, wordnet, and geonames}.
\newblock In \emph{Proceedings of the Fifteenth International Semantic Web
  Conference (ISWC2016)}, pages 177--185.

\bibitem[{Suchanek et~al.(2011)Suchanek, Abiteboul, and Senellart}]{paris}
Fabian~M. Suchanek, Serge Abiteboul, and Pierre Senellart. 2011.
\newblock \href {https://doi.org/10.14778/2078331.2078332} {Paris:
  probabilistic alignment of relations, instances, and schema}.
\newblock \emph{Proc. VLDB Endow.}, 5(3):157--168.

\bibitem[{Sun et~al.(2017)Sun, Hu, and Li}]{jape}
Zequn Sun, Wei Hu, and Chengkai Li. 2017.
\newblock \href {https://doi.org/10.1007/978-3-319-68288-4_37} {Cross-lingual
  entity alignment via joint attribute-preserving embedding}.
\newblock In \emph{The Semantic Web--ISWC 2017: 16th International Semantic Web
  Conference, Vienna, Austria, October 21--25, 2017, Proceedings, Part I 16},
  pages 628--644. Springer.

\bibitem[{Sun et~al.(2018)Sun, Hu, Zhang, and Qu}]{bootea}
Zequn Sun, Wei Hu, Qingheng Zhang, and Yuzhong Qu. 2018.
\newblock \href {https://doi.org/10.24963/ijcai.2018/611} {Bootstrapping entity
  alignment with knowledge graph embedding}.
\newblock In \emph{Proceedings of the Twenty-Seventh International Joint
  Conference on Artificial Intelligence, {IJCAI-18}}, pages 4396--4402.
  International Joint Conferences on Artificial Intelligence Organization.

\bibitem[{Sun et~al.(2020{\natexlab{a}})Sun, Wang, Hu, Chen, Dai, Zhang, and
  Qu}]{AliNet}
Zequn Sun, Chengming Wang, Wei Hu, Muhao Chen, Jian Dai, Wei Zhang, and Yuzhong
  Qu. 2020{\natexlab{a}}.
\newblock \href {https://doi.org/10.1609/aaai.v34i01.5354} {Knowledge graph
  alignment network with gated multi-hop neighborhood aggregation}.
\newblock \emph{Proceedings of the AAAI Conference on Artificial Intelligence},
  34(01):222--229.

\bibitem[{Sun et~al.(2020{\natexlab{b}})Sun, Zhang, Hu, Wang, Chen, Akrami, and
  Li}]{openea}
Zequn Sun, Qingheng Zhang, Wei Hu, Chengming Wang, Muhao Chen, Farahnaz Akrami,
  and Chengkai Li. 2020{\natexlab{b}}.
\newblock \href {https://api.semanticscholar.org/CorpusID:212737039} {A
  benchmarking study of embedding-based entity alignment for knowledge graphs}.
\newblock \emph{Proceedings of the VLDB Endowment}, 13:2326 -- 2340.

\bibitem[{Tang et~al.(2023)Tang, Zhao, and Li}]{fgwea}
Jianheng Tang, Kangfei Zhao, and Jia Li. 2023.
\newblock \href {https://doi.org/10.18653/v1/2023.findings-acl.205} {A fused
  {G}romov-{W}asserstein framework for unsupervised knowledge graph entity
  alignment}.
\newblock In \emph{Findings of the Association for Computational Linguistics:
  ACL 2023}, pages 3320--3334, Toronto, Canada. Association for Computational
  Linguistics.

\bibitem[{Tang et~al.(2020)Tang, Zhang, Chen, Yang, Chen, and Li}]{bertint}
Xiaobin Tang, Jing Zhang, Bo~Chen, Yang Yang, Hong Chen, and Cuiping Li. 2020.
\newblock \href {https://doi.org/10.24963/ijcai.2020/439} {{BERT-INT}:a
  bert-based interaction model for knowledge graph alignment}.
\newblock In \emph{Proceedings of the Twenty-Ninth International Joint
  Conference on Artificial Intelligence, {IJCAI-20}}, pages 3174--3180.
  International Joint Conferences on Artificial Intelligence Organization.
\newblock Main track.

\bibitem[{Vrande\v{c}i\'{c} and Kr\"{o}tzsch(2014)}]{wikidata}
Denny Vrande\v{c}i\'{c} and Markus Kr\"{o}tzsch. 2014.
\newblock \href {https://doi.org/10.1145/2629489} {Wikidata: a free
  collaborative knowledgebase}.
\newblock \emph{Commun. ACM}, 57(10):78--85.

\bibitem[{Wang et~al.(2018)Wang, Lv, Lan, and Zhang}]{gcnalign}
Zhichun Wang, Qingsong Lv, Xiaohan Lan, and Yu~Zhang. 2018.
\newblock \href {https://doi.org/10.18653/v1/D18-1032} {Cross-lingual knowledge
  graph alignment via graph convolutional networks}.
\newblock In \emph{Proceedings of the 2018 Conference on Empirical Methods in
  Natural Language Processing}, pages 349--357, Brussels, Belgium. Association
  for Computational Linguistics.

\bibitem[{Wu et~al.(2019{\natexlab{a}})Wu, Liu, Feng, Wang, Yan, and
  Zhao}]{rdgcn}
Yuting Wu, Xiao Liu, Yansong Feng, Zheng Wang, Rui Yan, and Dongyan Zhao.
  2019{\natexlab{a}}.
\newblock \href {https://doi.org/10.24963/ijcai.2019/733} {Relation-aware
  entity alignment for heterogeneous knowledge graphs}.
\newblock In \emph{Proceedings of the Twenty-Eighth International Joint
  Conference on Artificial Intelligence, {IJCAI-19}}, pages 5278--5284.
  International Joint Conferences on Artificial Intelligence Organization.

\bibitem[{Wu et~al.(2019{\natexlab{b}})Wu, Liu, Feng, Wang, and Zhao}]{hgcn}
Yuting Wu, Xiao Liu, Yansong Feng, Zheng Wang, and Dongyan Zhao.
  2019{\natexlab{b}}.
\newblock \href {https://doi.org/10.18653/v1/D19-1023} {Jointly learning entity
  and relation representations for entity alignment}.
\newblock In \emph{Proceedings of the 2019 Conference on Empirical Methods in
  Natural Language Processing and the 9th International Joint Conference on
  Natural Language Processing (EMNLP-IJCNLP)}, pages 240--249, Hong Kong,
  China. Association for Computational Linguistics.

\bibitem[{Wu et~al.(2020)Wu, Liu, Feng, Wang, and Zhao}]{nmn}
Yuting Wu, Xiao Liu, Yansong Feng, Zheng Wang, and Dongyan Zhao. 2020.
\newblock \href {https://doi.org/10.18653/v1/2020.acl-main.578} {Neighborhood
  matching network for entity alignment}.
\newblock In \emph{Proceedings of the 58th Annual Meeting of the Association
  for Computational Linguistics}, pages 6477--6487, Online. Association for
  Computational Linguistics.

\bibitem[{Xu et~al.(2019)Xu, Wang, Yu, Feng, Song, Wang, and Yu}]{gmnn}
Kun Xu, Liwei Wang, Mo~Yu, Yansong Feng, Yan Song, Zhiguo Wang, and Dong Yu.
  2019.
\newblock \href {https://doi.org/10.18653/v1/P19-1304} {Cross-lingual knowledge
  graph alignment via graph matching neural network}.
\newblock In \emph{Proceedings of the 57th Annual Meeting of the Association
  for Computational Linguistics}, pages 3156--3161, Florence, Italy.
  Association for Computational Linguistics.

\bibitem[{Yang et~al.(2019)Yang, Zou, Shi, Lu, Lin, and Sun}]{hman}
Hsiu-Wei Yang, Yanyan Zou, Peng Shi, Wei Lu, Jimmy Lin, and Xu~Sun. 2019.
\newblock \href {https://doi.org/10.18653/v1/D19-1451} {Aligning cross-lingual
  entities with multi-aspect information}.
\newblock In \emph{Proceedings of the 2019 Conference on Empirical Methods in
  Natural Language Processing and the 9th International Joint Conference on
  Natural Language Processing (EMNLP-IJCNLP)}, pages 4431--4441, Hong Kong,
  China. Association for Computational Linguistics.

\bibitem[{Yang et~al.(2024)Yang, Chen, Wang, Yang, Wang, and Liu}]{LLMEA}
Linyao Yang, Hongyang Chen, Xiao Wang, Jing Yang, Fei-Yue Wang, and Han Liu.
  2024.
\newblock \href {http://arxiv.org/abs/2401.16960} {Two heads are better than
  one: Integrating knowledge from knowledge graphs and large language models
  for entity alignment}.

\bibitem[{Zeng et~al.(2022)Zeng, Dong, Hou, Cao, Hu, Yu, Lv, Cao, Wang, Liu,
  Huang, Feng, Wan, Li, and Feng}]{iclea}
Kaisheng Zeng, Zhenhao Dong, Lei Hou, Yixin Cao, Minghao Hu, Jifan Yu, Xin Lv,
  Lei Cao, Xin Wang, Haozhuang Liu, Yi~Huang, Junlan Feng, Jing Wan, Juanzi Li,
  and Ling Feng. 2022.
\newblock \href {https://doi.org/10.1145/3511808.3557364} {Interactive
  contrastive learning for self-supervised entity alignment}.
\newblock In \emph{Proceedings of the 31st ACM International Conference on
  Information \& Knowledge Management}, CIKM '22, pages 2465--2475, New York,
  NY, USA. Association for Computing Machinery.

\bibitem[{Zeng et~al.(2020)Zeng, Zhao, Tang, and Lin}]{CEA-ICDE2020}
Weixin Zeng, Xiang Zhao, Jiuyang Tang, and Xuemin Lin. 2020.
\newblock \href {https://doi.org/10.1109/ICDE48307.2020.00191} {Collective
  entity alignment via adaptive features}.
\newblock In \emph{2020 IEEE 36th International Conference on Data Engineering
  (ICDE)}, pages 1870--1873.

\bibitem[{Zhang et~al.(2019)Zhang, Sun, Hu, Chen, Guo, and Qu}]{multike}
Qingheng Zhang, Zequn Sun, Wei Hu, Muhao Chen, Lingbing Guo, and Yuzhong Qu.
  2019.
\newblock \href {https://doi.org/10.24963/ijcai.2019/754} {Multi-view knowledge
  graph embedding for entity alignment}.
\newblock In \emph{Proceedings of the Twenty-Eighth International Joint
  Conference on Artificial Intelligence, {IJCAI-19}}, pages 5429--5435.
  International Joint Conferences on Artificial Intelligence Organization.

\bibitem[{Zhang et~al.(2023)Zhang, Su, Trisedya, Zhao, Yang, Cheng, and
  Qi}]{autoalign}
Rui Zhang, Yixin Su, Bayu~Distiawan Trisedya, Xiaoyan Zhao, Min Yang, Hong
  Cheng, and Jianzhong Qi. 2023.
\newblock \href {https://doi.org/10.1109/TKDE.2023.3325484} {Autoalign: Fully
  automatic and effective knowledge graph alignment enabled by large language
  models}.
\newblock \emph{IEEE Transactions on Knowledge and Data Engineering}, pages
  1--14.

\bibitem[{Zhang et~al.(2020)Zhang, Liu, Chen, Chen, Liu, Xiang, and
  Zheng}]{medbbk9k}
Ziheng Zhang, Hualuo Liu, Jiaoyan Chen, Xi~Chen, Bo~Liu, YueJia Xiang, and
  Yefeng Zheng. 2020.
\newblock \href {https://doi.org/10.18653/v1/2020.coling-industry.17} {An
  industry evaluation of embedding-based entity alignment}.
\newblock In \emph{Proceedings of the 28th International Conference on
  Computational Linguistics: Industry Track}, pages 179--189, Online.
  International Committee on Computational Linguistics.

\bibitem[{Zhao et~al.(2023)Zhao, Wu, Cai, Zhang, Zhang, and Yuan}]{tea}
Yu~Zhao, Yike Wu, Xiangrui Cai, Ying Zhang, Haiwei Zhang, and Xiaojie Yuan.
  2023.
\newblock \href {https://doi.org/10.18653/v1/2023.findings-acl.559} {From
  alignment to entailment: A unified textual entailment framework for entity
  alignment}.
\newblock In \emph{Findings of the Association for Computational Linguistics:
  ACL 2023}, pages 8795--8806, Toronto, Canada. Association for Computational
  Linguistics.

\bibitem[{Zhong et~al.(2022)Zhong, Zhang, Fan, and Dou}]{SDEA}
Ziyue Zhong, Meihui Zhang, Ju~Fan, and Chenxiao Dou. 2022.
\newblock \href {https://doi.org/10.1109/ICDE53745.2022.00205} {Semantics
  driven embedding learning for effective entity alignment}.
\newblock In \emph{2022 IEEE 38th International Conference on Data Engineering
  (ICDE)}, pages 2127--2140.

\bibitem[{Zhu et~al.(2017)Zhu, Xie, Liu, and Sun}]{zhu2017iterative}
Hao Zhu, Ruobing Xie, Zhiyuan Liu, and Maosong Sun. 2017.
\newblock \href {https://doi.org/10.24963/ijcai.2017/595} {Iterative entity
  alignment via joint knowledge embeddings}.
\newblock In \emph{Proceedings of the Twenty-Sixth International Joint
  Conference on Artificial Intelligence, {IJCAI-17}}, pages 4258--4264.

\bibitem[{Zhu et~al.(2019)Zhu, Zhou, Wu, Tan, and Guo}]{naea2019}
Qiannan Zhu, Xiaofei Zhou, Jia Wu, Jianlong Tan, and Li~Guo. 2019.
\newblock \href {https://doi.org/10.24963/ijcai.2019/269} {Neighborhood-aware
  attentional representation for multilingual knowledge graphs}.
\newblock In \emph{Proceedings of the Twenty-Eighth International Joint
  Conference on Artificial Intelligence, {IJCAI-19}}, pages 1943--1949.
  International Joint Conferences on Artificial Intelligence Organization.

\end{thebibliography}
\bibliographystyle{acl_natbib}

\end{document}